\documentclass[dvipsnames]{bmvc2k}
\usepackage{url}
\usepackage{color}
\usepackage{multirow}
\usepackage{graphicx}
\usepackage{float} 
\usepackage{tabularx}
\usepackage{multirow}
\usepackage{makecell}
\usepackage{booktabs}
\usepackage{graphicx}

\usepackage{subcaption}
\usepackage{ragged2e} 
\usepackage{epigraph}
\usepackage{placeins}
\usepackage{caption}
\usepackage{amsfonts}
\usepackage{tgheros}
\usepackage{wrapfig}
\usepackage{xr-hyper}   
\definecolor{customblue}{HTML}{091A61}

\usepackage{hyperref}
\hypersetup{
  colorlinks=true,
  linkcolor=MidnightBlue,
  citecolor=OliveGreen,
  urlcolor=BrickRed
}

\newcommand{\rev}[1]{\textcolor{black}{#1}}     

\title{\textit{What Can We Learn from Harry Potter?} \\An Exploratory Study of\\ Visual Representation Learning from Atypical Videos}

\addauthor{Qiyue Sun}{julysun@mail.sdu.edu.cn}{1,2}
\addauthor{Qiming Huang}{qxh366@student.bham.ac.uk}{1}
\addauthor{Yang Yang}{yyang@sdu.edu.cn}{2}
\addauthor{Hongjun Wang}{hjw@sdu.edu.cn}{2}

\addauthor{Jianbo Jiao}{j.jiao@bham.ac.uk}{1}
\addinstitution{The MIx Group, School of Computer Science, University of Birmingham, Birmingham, UK \vspace{1.3cm}}
\addinstitution{School of Information Science and Engineering, Shandong University, Qingdao, China}

\runninghead{Sun et al.}{What Can We Learn from Harry Potter?}

\def\eg{\emph{e.g}\bmvaOneDot}

\def\etal{\emph{et al}\bmvaOneDot}
\def\etc{\emph{etc}\bmvaOneDot}

\begin{document}
\maketitle
\vspace{-1em}
\vspace{-3mm}
\begin{abstract}
Humans usually show exceptional generalisation and discovery ability in the open world, when being shown uncommon new concepts. Whereas most existing studies in the literature focus on common typical data from closed sets, open-world novel discovery is under-explored in videos.
In this paper, we are interested in asking: \textit{What if atypical unusual videos are exposed in the learning process?}
To this end, we collect a new video dataset consisting of various types of unusual atypical data (\eg sci-fi, animation, \etc). To study how such atypical data may benefit open-world learning, we feed them into the model training process for representation learning.
Focusing on three key tasks in open-world learning: out-of-distribution (OOD) detection, novel category discovery (NCD), and zero-shot action recognition (ZSAR), we found that even straightforward learning approaches with atypical data consistently improve performance across various settings. Furthermore, we found that increasing the categorical diversity of the atypical samples further boosts OOD detection performance. Additionally, in the NCD task, using a smaller yet more semantically diverse set of atypical samples leads to better performance compared to using a larger but more typical dataset. In the ZSAR setting, the semantic diversity of atypical videos helps the model generalise better to unseen action classes.
These observations in our extensive experimental evaluations reveal the benefits of atypical videos for visual representation learning in the open world, together with the newly proposed dataset, encouraging further studies in this direction.
The project page is at: \url{https://julysun98.github.io/atypical_dataset}.
\end{abstract}

\epigraph{\makebox[0.9\textwidth][c]{\hspace{-12mm}``Discovery commences with the awareness of anomaly''}}{\textit{--- Thomas S. Kuhn}}

\vspace{-1em}
\section{Introduction}
\label{sec:intro}
\begin{figure*}[htb]
\begin{center}
\includegraphics[scale=0.3]{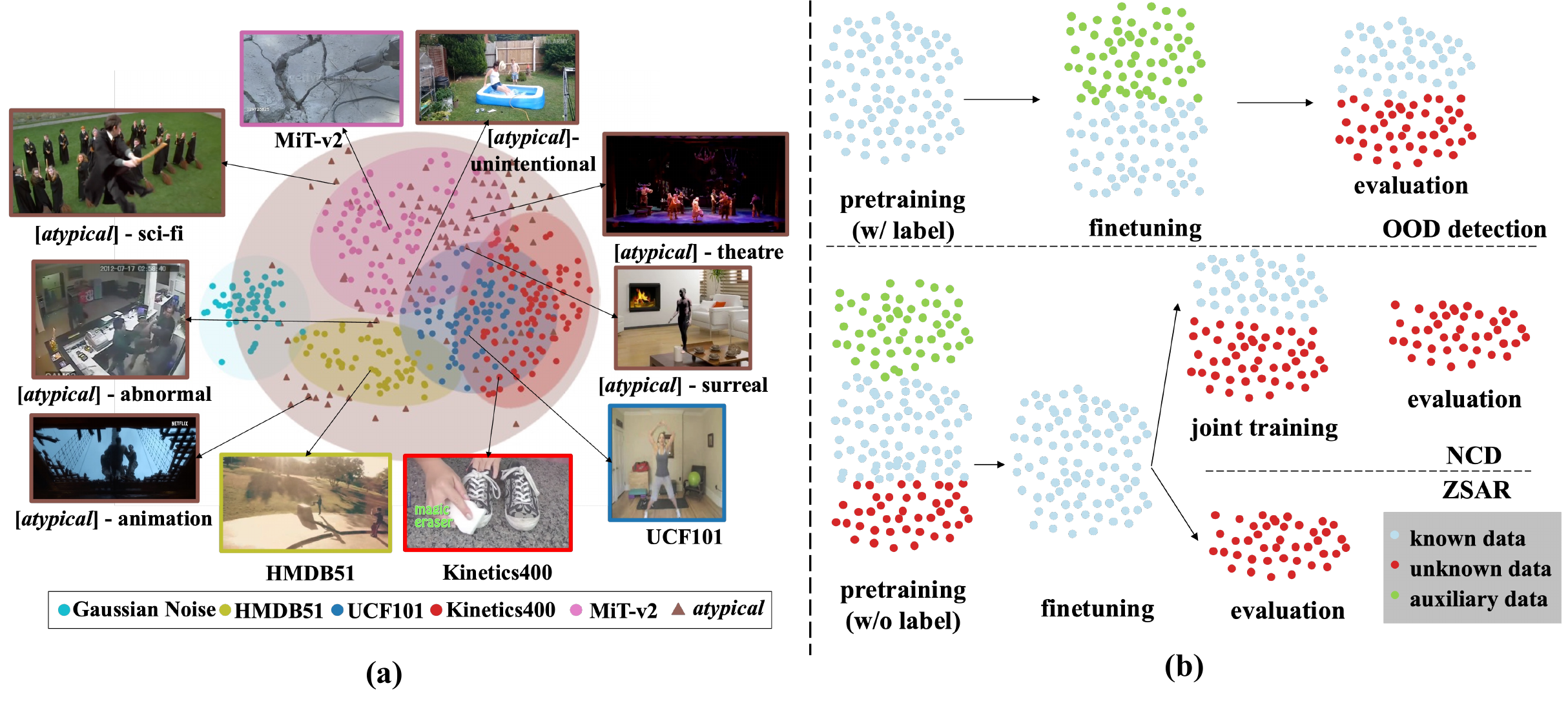}
\end{center}
\caption{Illustration of open-world data and tasks. (a) Comparison of open-world and closed-world data distributions. Commonly used datasets such as Kinetics-400~\citep{kay2017kinetics}, UCF101~\citep{soomro2012ucf101}, HMDB51~\citep{kuehne2011hmdb}, and MiT-v2~\citep{monfort2019moments} represent more concentrated distributions and are considered closed-world. In contrast, our proposed \textit{atypical} dataset captures a broader, more diverse feature space, better reflecting the open-world setting. (b) Overview of open-world tasks: OOD detection identifies out-of-distribution (unknown) data from known categories; novel category discovery (NCD) clusters the unknown data to reveal new classes; and zero-shot action recognition (ZSAR) further classifies these new categories using semantic information. These tasks form a natural progression, with increasing difficulty and reliance on model generalisation.}
\label{fig:intro}
\vspace{-1em}
\end{figure*}

Human cognition excels at generalising from limited information and discovering new concepts in dynamic and unpredictable environments~\citep{lieder2020resource,saxe2021if}. This ability to adapt to unfamiliar stimuli in an open world contrasts with the limitations faced by current machine learning models~\citep{Heigold_2023_ICCV}, especially in the field of video understanding. Current models operate mainly in closed hypothetical environments where all possible categories are predefined during training, which limits their ability to handle the variety of unpredictable scenarios often encountered in real-world applications~\citep{Zhou_2021_CVPR,kejriwal2024challenges}. The question remains whether models can be enhanced to navigate the open world with the same adaptability as human cognition.

Previous advancements in video understanding have largely focused on closed-set environments, where the model is trained and tested on well-curated~\citep{zhu2022towards}, typical datasets such as UCF101~\citep{soomro2012ucf101}, Kinetics-400~\citep{kay2017kinetics}, and HMDB51~\citep{kuehne2011hmdb}.  Although these models perform well within known distributions, they encounter significant difficulties when exposed to out-of-distribution (OOD) new data~\citep{acsintoae2022ubnormal,rame2022fishr}, thereby limiting their applicability to open-world environments where new and unknown categories frequently emerge~\citep{chen2023wdiscood,ming2022poem}. There are also ways to use generative modelling, such as GANs~\citep{kong2021opengan,grcic2020dense} to generate virtual data or virtual features to help with OOD detection~\citep{du2022towards}, However, the data generated using this approach may lack adequate diversity and differ somewhat from real-world data distributions. Existing datasets, despite being useful benchmarks, do not encourage models to generalise beyond the constraints of the training distribution~\citep{zhang2021understanding}. As a result, the challenge of detecting and adapting to novel instances in the open world remains an underdeveloped area in video representation learning.

The above-mentioned observations and limitations in current closed-set studies \rev{motivated us to ask} a question: \textit{Would that help the open-world understanding and learning if atypical and uncommon videos are introduced to the model?} By exposing models to data that lies outside the typical distribution, we argue that it may lead to a more robust capacity for open-world learning. Addressing this question necessitates a reconsideration of traditional video classification datasets and opens the possibility of utilising more diverse and atypical data during training.

\textit{Atypical} data, characterised by its departure from common real-world categories, offers a unique opportunity to challenge and enhance model generalisation. Unlike conventional datasets, which primarily consist of daily activities, atypical data refers to a wide range of unusual and outlier scenarios, including those found in science fiction, animation, and anomalous real-world situations. These atypical samples present a broader spectrum of visual content, allowing models to learn from examples that deviate from the norm~\citep{rame2022fishr}. We anticipate that incorporating such data during training will enable the model to handle open-world environments more effectively.
To this end, in this paper, we present the first atypical dataset to support the study in this direction.

In deep learning, particularly in open-world environments, models often encounter data that deviates from the distribution of their training data~\citep{chen2023wdiscood}. To systematically evaluate the effectiveness of learning with atypical data, we leverage three fundamental open-world tasks: OOD detection~\citep{hendrycks2016baseline}, NCD~\citep{han2021autonovel} and ZSAR. These tasks, while distinct, form a progressive framework for open-world video understanding. OOD detection serves as the first step, where the model must identify whether an input sample deviates from the training distribution and should be treated as novel. Upon detecting such unfamiliar inputs, NCD aims to organise these unknown samples into coherent, previously unseen categories, often without any labels~\citep{yang2024generalized}. Building on this, ZSAR takes one step further by asking the model not only to detect or group novel samples, but to directly recognise instances of unseen action classes using semantic knowledge transferred from seen categories. Together, these tasks represent a continuum: from detecting the unknown (OOD), to structuring the unknown (NCD), and finally to understanding the unknown (ZSAR). This progression captures increasingly sophisticated aspects of generalisation in open-world scenarios and allows for a comprehensive evaluation of models trained with atypical data. An illustration is shown in Figure~\ref{fig:intro}.

To incorporate \textit{atypical} data during training for open-world tasks, we adopt auxiliary data to enhance learning~\citep{hendrycks2018deep,papadopoulos2021outlier,zhu2023diversified,zhang2023mixture}. For OOD detection tasks, the core idea is to use auxiliary outliers during training, enabling the model to more effectively distinguish between in-distribution (ID) samples and OOD samples~\citep{ming2022poem}. In NCD, introducing auxiliary data mirrors the human learning process -- exposure to diverse information, even subconsciously, helps the brain adapt and learn new knowledge when provided with some context. For ZSAR, atypical data can provide indirect exposure to semantically rich or stylistically diverse actions, which supports the model’s ability to recognise unseen categories via transferable representations. However, bridging the distribution gap between surrogate new data and unseen inputs remains challenging~\citep{zhu2023unleashing}, as it is hard to know the prior knowledge of potential OOD inputs that would be encountered at the inference stage~\citep{zhu2023diversified}. Our approach seeks to mitigate this by using a newly proposed diverse and atypical dataset during training, aiming to better equip models with the capability to handle a wide range of potential open-world scenarios. 

Extensive experiments show that incorporating atypical video data (\eg sci-fi, animation, abnormal, and unintentional) improves model performance across multiple open-world tasks, including OOD detection, NCD, and ZSAR. Our analyses suggest that the diversity of atypical samples is critical in this process. Models trained with more diverse atypical datasets exhibit greater robustness and superior performance in identifying new and unseen distributions due to the endless availability of new knowledge. These findings highlight the potential of atypical data for enhancing visual representation learning in open-world environments, pointing to promising avenues for further exploration in this direction.

\section{Related Work}
\label{relatedwork}

\vspace{-2mm}
\subsection{Video datasets}
Video datasets have been vital in advancing computer vision, particularly for recognising human behaviour through video analysis~\citep{kuehne2011hmdb,kay2017kinetics,soomro2012ucf101,wang2014action,miech19howto100m,chung2021haa500}. While numerous datasets have been developed to support this line of research, most focus on closed-set tasks, such as classifying and localising predefined human movements~\citep{poppe2010survey,kong2022human,sun2022human}. Although effective for benchmarking model performance on specific objectives, these datasets fall short in representing atypical scenarios—rare, extreme, or fictional events that are frequently encountered in real-world applications~\citep{acsintoae2022ubnormal}. Consequently, models trained on such datasets often struggle to generalise to the complex and diverse conditions of open-world environments. To address this limitation, we propose leveraging atypical video data, including footage from anomaly detection, unintended actions, synthetic media, theatrical performances, and fictional or animated content. We argue that such data is essential for open-world learning tasks in the video domain, as it exposes models to a broader spectrum of variability.
\vspace{-1.2em}
\subsection{Open-world learning}
Open-world learning~\citep{kong2021opengan,yang2022openood,vaze2021open} addresses the challenge of recognising and adapting to novel inputs beyond the training distribution. Traditional learning paradigms operate under the closed-world assumption, which restricts their ability to generalise to unseen scenarios~\citep{scheirer2011meta, bendale2016towards}. This limitation motivates the development of more adaptive approaches that can effectively handle the variability encountered in real-world environments.

\noindent \textbf{Out-of-distribution (OOD) detection}.
OOD detection aims to determine whether an input sample originates from a known in-distribution (ID) or from an out-of-distribution (OOD) source. Yang \etal~\citep{yang2022openood} introduced OpenOOD, a benchmark for evaluating OOD detection across diverse settings. Hsu \etal~\citep{hsu2020generalized} proposed Generalised ODIN, which enhances model robustness through input preprocessing and temperature scaling, without the need for external OOD data. In contrast, our work demonstrates that incorporating atypical datasets during training significantly boosts OOD detection performance.

\noindent \textbf{Novel category discovery}.
NCD extends OOD detection by identifying and grouping unknown categories within unlabelled data~\citep{hsu2020generalized,han2019learning}. Zhong \etal~\citep{zhong2021openmix} proposed OpenMix, which improves performance by mixing known and novel data during training. We also highlight NovelCraft~\citep{feeney2022novelcraft}, which introduces a new dataset specifically designed for novelty detection and discovery in open-world contexts. Unlike their dataset—based on visuals from the game Minecraft—our dataset better reflects the data distribution in real-world scenarios.

\noindent \textbf{Zero-shot action recognition}.
ZSAR tackles the problem of recognising action categories that are absent from the training set, typically by exploiting semantic relationships between known and unseen classes~\citep{zhu2018towards}. Common methods rely on external textual sources or vision-language models to enable cross-modal generalisation~\citep{wang2021actionclip}. However, many such methods are constrained by limited domain exposure and reliance on handcrafted prompts or taxonomies. Our dataset introduces a broader range of visual contexts, potentially enhancing generalisation in ZSAR.

\begin{figure*}[t]
\begin{center}
\includegraphics[width=1\textwidth]{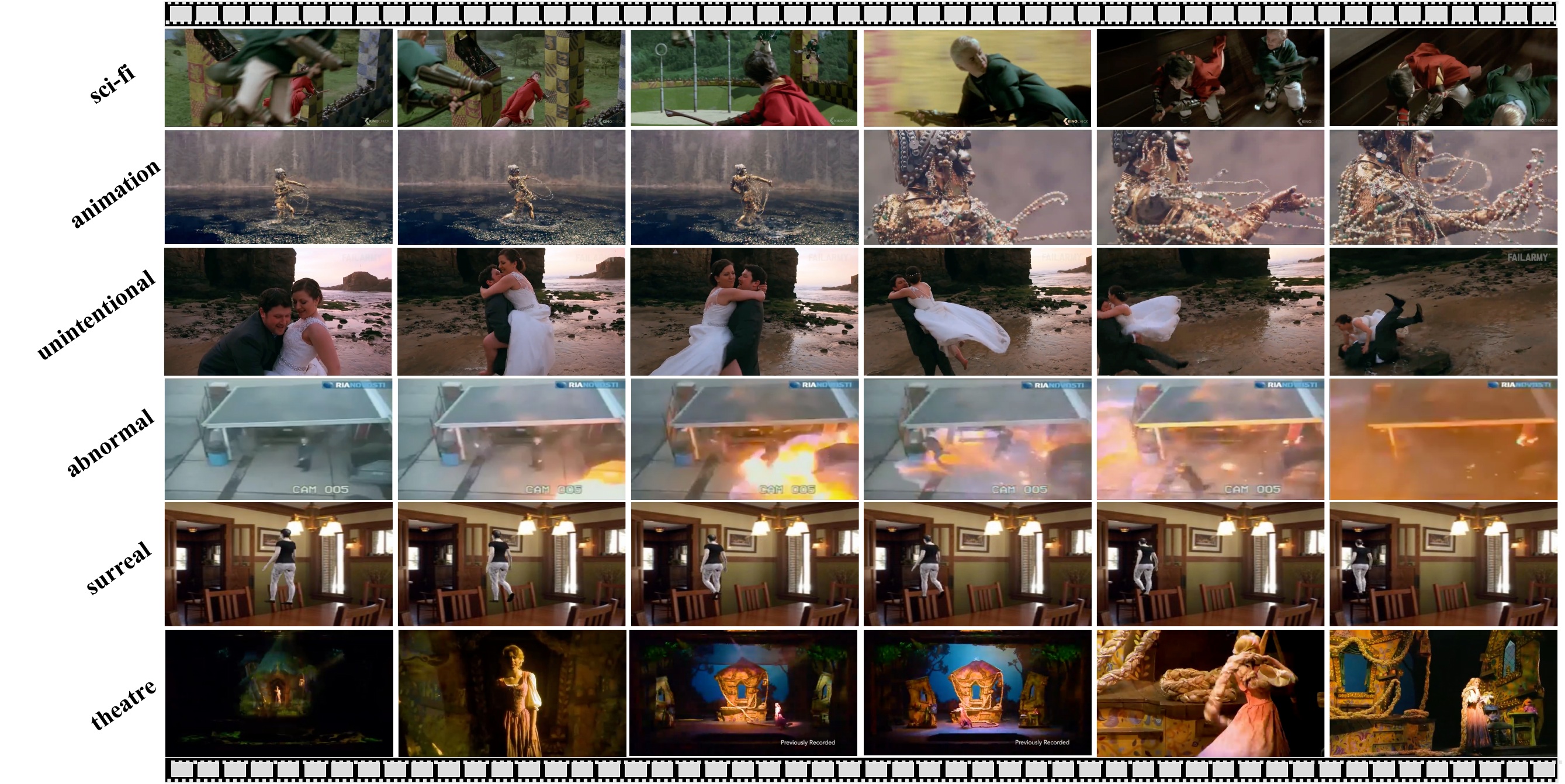}
\end{center}
\vspace{-1mm}
\caption{Representative examples from the proposed \textit{atypical} video dataset.}
\label{fig:illustration}
\end{figure*}

\section{The Atypical Dataset}
\label{atypical}
\vspace{-2mm}
Here we introduce, to the best of our knowledge, the first \textit{atypical} video dataset, consisting of various scenarios that are uncommon in real life. We then use this dataset for the following open-world learning study. Specifically, the dataset consists of 7,818 video clips sourced from prior datasets and YouTube. These videos depict abnormal, unintentional, and uncommon activities in the open world, as well as unreal videos such as sci-fi movies, animations, and synthetic scenes. Figure~\ref{fig:illustration} shows some examples of the \textit{atypical} dataset. Different from existing action classification and video understanding datasets, our \textit{atypical} data focuses on rare or uncommon video activities and even activities that do not occur in the real world. \rev{More examples can be found in in Appendix}~\ref{Appendix_A}.

\vspace{-2mm}
\subsection{Data sources and pre-processing}
\vspace{-0.5em}
The \textit{atypical} video dataset comprises multiple sources, each significantly deviating from typical behavioural patterns or conventional visual content commonly observed in real-world videos. To construct the dataset, we apply the following processing procedures.

\vspace{-3mm}
\paragraph{Unintentional, Abnormal, and Surreal.} For these categories, we sampled a subset of videos from existing datasets and segmented them into clips of 5–10 seconds. Unintentional behaviour clips were sourced from the Oops Dataset~\citep{epstein2020oops}, while abnormal scenes were taken from established anomaly detection datasets such as UCSD Ped2~\citep{Mahadevan2010anomaly}, CUHK Avenue~\citep{lu2013abnormal}, and UCF-Crime~\citep{sultani2018real}. Surreal content was drawn from the Surreal dataset~\citep{varol17_surreal}, which features photorealistic synthetic human renderings. These subsets collectively capture accidental actions, anomalous behaviours, and artificial human interactions, offering distinct deviations from typical real-world action patterns.

\vspace{-3mm}
\paragraph{Sci-fi, Animation, and Theatre.}
For these categories, we collected trailers and short clips from YouTube, segmenting them into 2–6 second intervals to maximise scene diversity. Manual filtering was applied to remove irrelevant content (\eg non-sci-fi scenes). Sci-fi clips were selected from live-action film trailers that depict futuristic or supernatural themes. Animated content was obtained from recent cinematic trailers that combine imaginative and realistic visual styles. Theatre clips consist of exaggerated, stylised stage performances curated from online videos. Each subset provides visually and semantically distinct scenarios not typically encountered in conventional video datasets.

\subsection{Dataset statistics}
\begin{wrapfigure}{r}{0.4\textwidth}
    \vspace{-3em}
    \centering
    \includegraphics[width=\linewidth, trim=55em 0 35em 3em, clip]{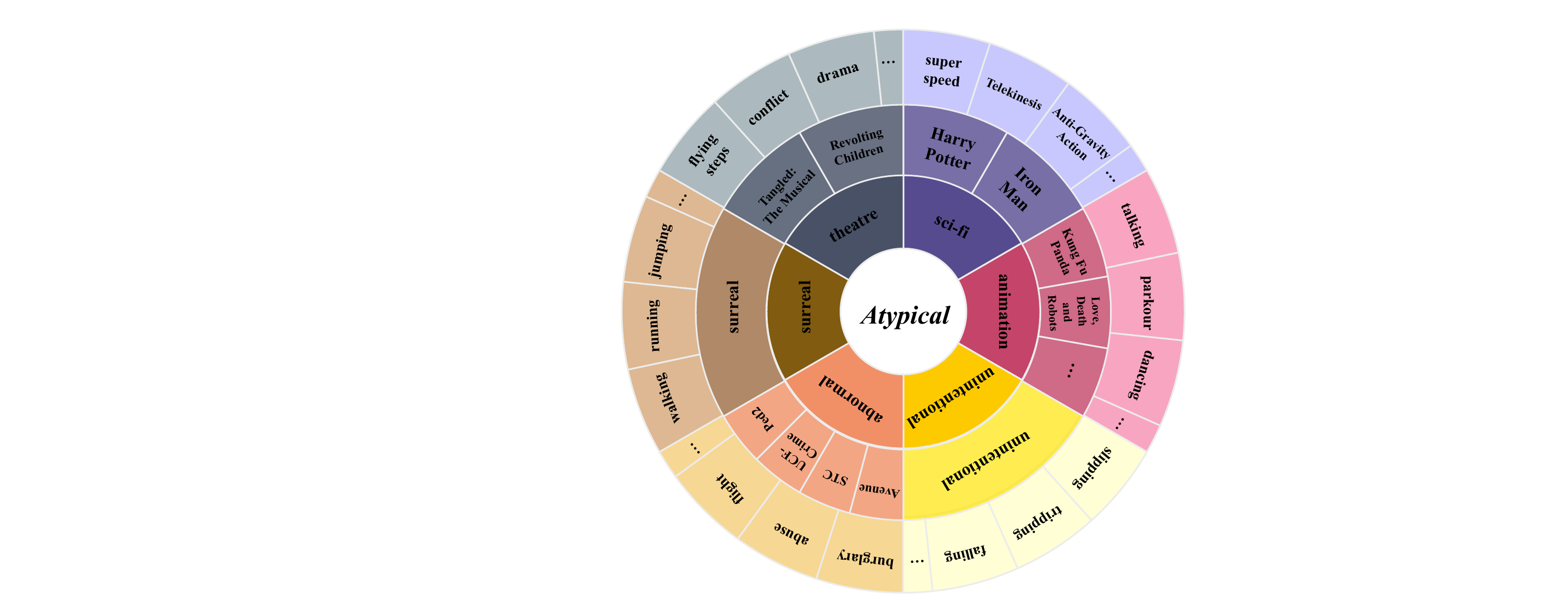}
    \vspace{-3mm}
    \caption{Composition of the introduced \textit{atypical} dataset.}
    \label{fig:stat}
    \vspace{-1.2em}
\end{wrapfigure}

\vspace{-0.5em}
Existing publicly available datasets primarily focus on common human actions and activities~\citep{soomro2012ucf101,kuehne2011hmdb,monfort2019moments,kay2017kinetics}. In contrast, our dataset introduces a broader spectrum of complex and diverse scenarios. As summarised in Table~\ref{tab:atypical_stat}, we categorised the dataset based on its source and content. Our \textit{atypical} dataset covers a broad range of scenes, subjects, actions, and other visual elements that are either rare or exhibit substantial semantic or visual deviation from those found in conventional, systematically curated datasets. This diversity is further illustrated in Figure~\ref{fig:stat}, which demonstrates the wide spectrum of unconventional behaviours captured across the atypical samples.


\begin{table}
\caption{Statistical details of the proposed \textit{atypical} video dataset.}
\small
\centering
\vspace{1em}
\begin{tabular}{lccl}
\toprule
\multicolumn{1}{l}{\textbf{Category}} & \multicolumn{1}{c}{\textbf{\# Videos}} & \textbf{Avg duration} & \textbf{Data Source} \\ \hline
Sci-fi & 1,119 & 4.00s & YouTube film trailers \\
Animation & 1,058 & 4.04s & YouTube animated trailers \\ 
Unintentional & 2,835 & 9.77s & Oops Dataset~\citep{epstein2020oops} \\
Abnormal & 1,103 & 7.53s & \makecell[l]{Ped2~\citep{Mahadevan2010anomaly}, CUHK Avenue~\citep{lu2013abnormal}, UCF-Crime~\citep{sultani2018real}} \\
Surreal & 1,024 & 3.18s & Surreal dataset~\cite{varol17_surreal} \\
Theatre & 679 & 4.81s & YouTube stage performances \\
\midrule
\textit{atypical} & 7,818 & 5.70s & Diverse sources \\
\bottomrule
\end{tabular}
\vspace{-1em}
\label{tab:atypical_stat}
\end{table}

\section{Effectiveness Analysis of Atypical Data}
\label{results}

\paragraph{Datasets.}
We follow standard dataset split settings for the OOD detection, NCD, and ZSAR tasks, as illustrated in Figure~\ref{fig:intro}(b). \rev{Brief descriptions and preliminaries of the OOD detection, NCD, and ZSAR tasks are provided in Appendix}~\ref{Appendix_tasks}. In all tasks, known data refers to in-distribution or labelled data used during training ($D_{in}$/$D_l$), while unknown data refers to out-of-distribution or unlabelled data ($D_{out}$/$D_u$). Auxiliary data ($D_{aux}$) without labels is introduced to support training. The datasets and split configurations are summarised in Table~\ref{tab:data_stat}. To ensure fair comparisons across datasets of varying sizes, we adopt a dynamic sampling strategy: for OOD detection, the size of the auxiliary data matches that of the ID set; for NCD and ZSAR, the amount of self-supervised training data is aligned with the evaluation dataset size. \rev{More details on data splits are provided in Appendix}~\ref{appendix_datasets_splits}
\vspace{-3mm}

\begin{wraptable}{r}{0.57\textwidth}
\centering
\footnotesize
\renewcommand{\arraystretch}{0.95}
\caption{Data splits for open world learning. The numbers 80, 40, 21, and 11 indicate the number of categories within each subset, respectively.}
\vspace{1em}
\label{tab:data_stat}
\begin{tabular}{l@{\hskip 3pt}c@{\hskip 3pt}c@{\hskip 3pt}c}

\toprule
\textbf{Task} & \textbf{Known} & \textbf{Unknown} & \textbf{Auxiliary} \\
\midrule
\makecell{OOD\\detection}    
& UCF101  
& \makecell{HMDB51\\MiT-v2\\{[}\textit{atypical}{]}\\-surreal\\{[}\textit{atypical}{]}\\-theatre}   
& \makecell{Gaussian\\Noise\\Diving48\\Kinetics400\\\textit{atypical}} \\
\midrule
\makecell{NCD /\\ZSAR}  
& \makecell{UCF\_80\\HMDB\_40}   
& \makecell{UCF\_21\\HMDB\_11} 
& \makecell{UCF101 / HMDB51\\Kinetics400\\\textit{atypical}}\\
\bottomrule
\end{tabular}
\end{wraptable}

\paragraph{Evaluation metrics \& implementation details.}
We adopt standard evaluation metrics for each task. For OOD detection, we report FPR95, AUROC, and AUPR, following~\citep{hendrycks2018deep,yang2022openood,zhu2023diversified,ming2022poem}. For NCD, clustering performance is measured using accuracy (acc), normalised mutual information (nmi), and adjusted Rand index (ari)~\citep{vaze2022generalized,han2021autonovel,vaze2023no,hanautomatically}. For ZSAR, we follow~\citep{ni2022expanding,wang2021actionclip,chen2021elaborative} and evaluate Top-1 and Top-5 accuracy (\%) on the set of novel classes.
%
%
For each task, we follow established training protocols from prior work. For OOD detection, we adopt the outlier exposure strategy introduced by~\citep{hendrycks2018deep}\rev{, using both the classical ResNet3D-50~\citep{hara2018can} backbone and the ViT-based TimeSformer backbone~\citep{gberta_2021_ICML}.} For NCD, we follow the AutoNovel framework~\citep{han2021autonovel}, incorporating self-supervised and joint training strategies\rev{, and employ ResNet3D-50~\citep{hara2018can} as the backbone.} For ZSAR, we build upon ActionCLIP~\citep{wang2021actionclip}, combining contrastive pre-training with supervised learning\rev{, using a CLIP-based backbone~\citep{radford2021learning}. Full training schedules and hyperparameter settings are presented in Appendix}~\ref{Appendix_Implementation}.

\vspace{-3mm}
\subsection{OOD detection}
For the OOD detection task, we explore the effect of introducing different auxiliary datasets during the training stage. Evaluation is conducted on both standard OOD benchmarks (HMDB51~\citep{kuehne2011hmdb}, MiT-v2~\citep{monfort2019moments}) and more challenging atypical distributions (\textit{atypical}-surreal, \textit{atypical}-theatre). The performance across all test sets is summarised in \rev{Table~\ref{tab:OOD_main_CR}}. Auxiliary datasets with limited semantic content, such as Gaussian noise and Diving48~\citep{li2018resound}, yield minimal improvements. In contrast, both Kinetics-400~\citep{kay2017kinetics} and our proposed \textit{atypical} dataset lead to notable gains, with \textit{atypical} achieving the best overall results across all metrics (FPR95, AUROC, AUPR). To validate the generality of these findings, we further conduct experiments using a ViT-based backbone~\citep{gberta_2021_ICML}, which exhibit consistent performance trends. \rev{ViT-based evaluations are included in Appendix}~\ref{Appendix_OOD}.
\vspace{-2mm}
%
\begin{table}[h]
\centering
\small
\caption{\rev{OOD detection performance (FPR95 ↓, AUROC ↑, AUPR ↑) using different auxiliary datasets during training with a ResNet3D-50 backbone. All models are evaluated on four testsets: HMDB51, MiT-v2, [\textit{atypical}]-surreal, and [\textit{atypical}]-theatre. \textit{Baseline}: training w/o auxiliary data. Best results are shown in \textbf{bold}.}}

\vspace{1em}
\setlength{\tabcolsep}{5pt}
\begin{tabular}{l|l|ccccc}
\toprule
\textbf{Dataset} & \textbf{Metric} & Baseline & +Gaussian & +Diving48 & +K400 & +\textit{Atypical} \\
\midrule
\multirow{3}{*}{HMDB51~\cite{kuehne2011hmdb}}
& FPR95 ↓ & 77.08 & 81.11 & 81.14 & 75.52 & \textbf{73.07} \\
& AUROC ↑ & 63.85 & 63.36 & 64.84 & 66.84 & \textbf{69.43} \\
& AUPR ↑  & 22.54 & 23.03 & 24.04 & 25.13 & \textbf{27.07} \\
\midrule
\multirow{3}{*}{MiT-v2~\cite{monfort2019moments}}
& FPR95 ↓ & 77.73 & 77.51 & 80.87 & 67.72 & \textbf{66.62} \\
& AUROC ↑ & 64.94 & 65.14 & 65.46 & 72.53 & \textbf{74.01} \\
& AUPR ↑  & 23.76 & 24.12 & 27.24 & 30.86 & \textbf{32.59} \\
\midrule
\multirow{3}{*}{[\textit{atypical}]-surreal}
& FPR95 ↓ & 62.22 & 67.37 & 73.14 & 53.66 & \textbf{49.99} \\
& AUROC ↑ & 75.50 & 73.32 & 69.13 & 79.78 & \textbf{82.03} \\
& AUPR ↑  & 33.42 & 32.26 & 25.71 & 39.13 & \textbf{43.19} \\
\midrule
\multirow{3}{*}{[\textit{atypical}]-theatre}
& FPR95 ↓ & 79.75 & 80.71 & 82.66 & 66.41 & \textbf{61.91} \\
& AUROC ↑ & 61.80 & 63.95 & 60.05 & 71.29 & \textbf{74.62} \\
& AUPR ↑  & 20.34 & 21.97 & 19.04 & 27.17 & \textbf{34.57} \\
\midrule
\midrule
\multirow{3}{*}
{Average}
&FPR95 ↓ & 74.20 & 76.68 & 79.45 & 65.83 & \textbf{62.90} \\
&AUROC ↑ & 66.52 & 66.44 & 64.87 & 72.61 & \textbf{75.02} \\
&AUPR ↑ & 25.02 & 25.35 & 24.01 & 30.62 & \textbf{34.36} \\
\bottomrule
\end{tabular}
\vspace{-1em}
\label{tab:OOD_main_CR}
\end{table}


\vspace{-2mm}
\subsection{Novel category discovery}
For the NCD task, we investigate the impact of different auxiliary datasets during the self-supervised pre-training stage. The baseline model is trained without auxiliary data. We compare Kinetics-400~\citep{kay2017kinetics}, the original dataset itself (UCF101~\citep{soomro2012ucf101}), our proposed \textit{atypical} dataset, and their combinations. As shown in Table~\ref{tab:ND_main}. When combined our \textit{atypical} data with UCF101~\citep{soomro2012ucf101}, it achieves the best overall results, surpassing configurations that include Kinetics-400~\citep{kay2017kinetics}. These findings highlight the effectiveness of \textit{atypical} in enhancing novel category discovery. \rev{Results on the HMDB\_11 split are provided in Appendix}~\ref{Appendix_NCD}.

\begin{table}[h]
\centering
\caption{NCD performance on UCF101 with different auxiliary datasets used for self-supervised pre-training. \textit{Baseline}: training w/o auxiliary data. UCF101 refers to using the original dataset itself as auxiliary data.}
\vspace{1em}
\small
\begin{tabular}{l|c|ccc|cc}
\toprule
\textbf{Metric} & Baseline &+UCF101 & +K400 &   +\textit{atypical} & +UCF101+K400 & +UCF101+\textit{atypical} \\
\midrule
acc ↑  & 19.85  & 21.86 & 20.81 & 21.99 & 26.57 & \textbf{27.88} \\
nmi ↑  & 0.2466 & 0.2966 & 0.2600 & 0.2737 & 0.3392 & \textbf{0.3511} \\
ari ↑  & 0.0797 & 0.1072 & 0.0772 &  0.0946 & 0.1240 & \textbf{0.1283} \\
\bottomrule
\end{tabular}
\vspace{-1.5em}
\label{tab:ND_main}
\end{table}

\subsection{Zero-shot action recognition}
Similar to the NCD task, we evaluate the impact of different auxiliary datasets on the performance of ZSAR. Results on the HMDB\_11 split are presented in Table~\ref{tab:ZSAR_main}, from which we can see that the results are consistent with our findings in NCD. \rev{Results on the UCF\_21 split are presented in Appendix}~\ref{Appendix_ZSAR}.

\vspace{-0.5em}

\begin{table}[h]
\centering
\caption{ZSAR performance on the HMDB\_11 split with different auxiliary datasets used for representation learning. \textit{Baseline}: training w/o auxiliary data.}
\vspace{1em}
\small
\resizebox{\textwidth}{!}{%
\begin{tabular}{l|c|ccc|cc}
\toprule
\textbf{Metric} & Baseline & +HMDB51 & +K400   & +\textit{atypical} & +HMDB51+K400 & +HMDB51+\textit{atypical} \\
\midrule
Top-1 ↑  & 57.40  & 57.02 & 58.23   & 58.30   & 61.31 & \textbf{62.96} \\
Top-5 ↑  & 94.97 & 96.62 & 96.84   & \textbf{97.45}  & 96.62 & 96.39 \\
\bottomrule
\end{tabular}
}
\vspace{-2em}
\label{tab:ZSAR_main}
\end{table}

\section{What Can We Learn from Atypical Videos?}
\label{discussion}

\subsection{\textit{Atypical} dataset exhibits semantic and distributional diversity}

Our \textit{atypical} dataset incorporates diverse scenarios such as sci-fi, animation, and unintentional footage, each contributing distinct visual and semantic characteristics. This variety is intended to enhance semantic diversity and better reflect the complexity of open-world video content. 
To quantify and compare the diversity across different datasets, we first extract feature representations using ResNet3D-50~\citep{hara2018can} and reduce their dimensionality with UMAP. 
\begin{wrapfigure}{r}{0.39\textwidth}
    \centering
    \includegraphics[width=\linewidth]{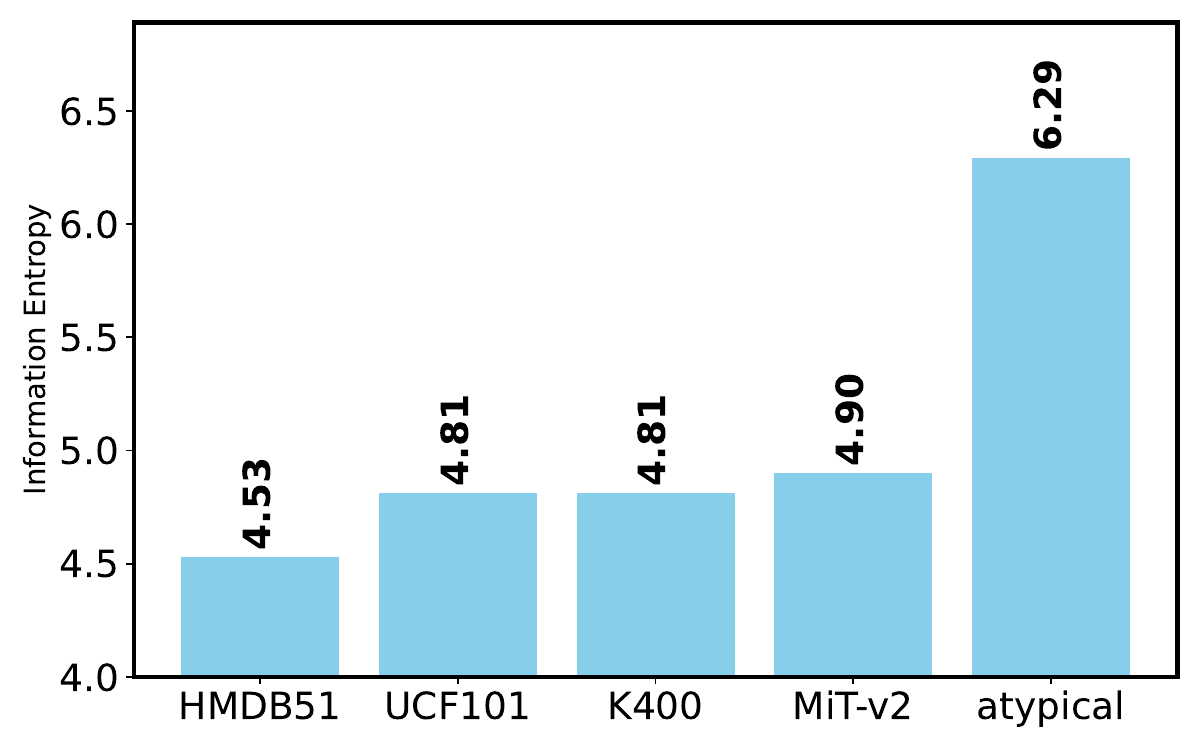}
    \vspace{-0.5em}
    \caption{Feature distributional entropy comparison between our \textit{atypical} dataset and typical datasets.}
    \label{fig:entropy}
    \vspace{-1em}
\end{wrapfigure}

The resulting low-dimensional embeddings are shown in Figure~\ref{fig:intro}(a). 
We then compute the information entropy of each dataset based on the distribution of its feature representations, using $H = -\sum_{i=1}^{N} p_i \log(p_i)$, where $p_i$ is the estimated probability of feature samples falling into discretised regions of the embedding space. The entropy values for different datasets are presented in Figure~\ref{fig:entropy}. A higher entropy indicates a more dispersed and diverse feature distribution, suggesting a broader coverage in the dataset's visual content.

\subsection{Semantic diversity helps open-world learning}
\begin{figure*}[ht]
    \begin{center} \includegraphics[width=1\textwidth, trim={0  0 0  0.5em}, clip]{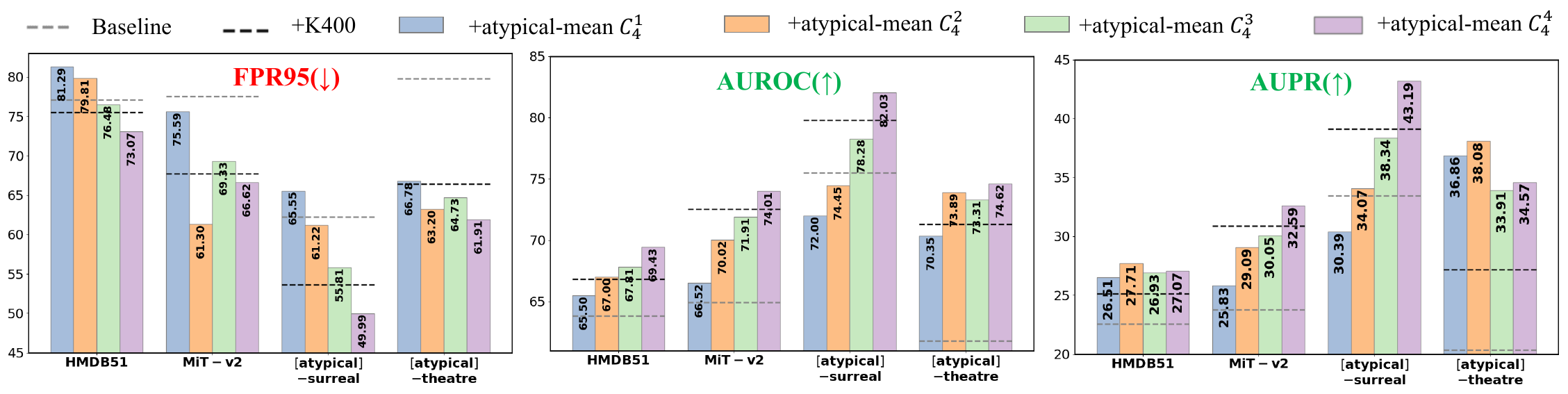}
    \end{center}
    \caption{Impact of introducing different numbers of semantic categories from the \textit{atypical} dataset on OOD task. Results are averaged over all combinations, where $C_4^i$ denotes the number of ways to select $i$ categories from four available \textit{atypical} types. Each subplot shows results on four test sets, where grouped bars represent different auxiliary data settings.}
    \label{fig:OOD_ablation}
    \vspace{-1em}
\end{figure*}

To examine the effect of semantic diversity in auxiliary data, we vary the number of semantic categories included from the \textit{atypical} dataset and evaluate OOD performance across multiple $D_{out}$ datasets. To ensure fair comparison, the total number of auxiliary samples is kept constant across all settings. As shown in Figure~\ref{fig:OOD_ablation}, introducing even a single \textit{atypical} category yields an improvement over the baseline trained solely on known data. Performance continues to increase as more categories are added, with average metrics improving steadily and standard deviation across test sets decreasing. These trends indicate that the diversity of semantic content contributes to more stable OOD. \rev{Detailed results for settings with one, two, and three semantic categories are provided in Appendix}~\ref{Appendix_diversity}.

\vspace{-0.5em}
\subsection{\textit{Atypical} data coverage helps open-world learning}
To further investigate the impact of content coverage in auxiliary data, we evaluate all pairwise combinations of \textit{atypical} subsets and their effects on OOD detection performance. Each subset captures distinct semantic content and visual style (\eg \textit{sci-fi}, \textit{animation}, \textit{unintentional}), resulting in varying levels of content coverage when combined. Table~\ref{tab:ood2} presents detailed results across four OOD test sets. For each metric, we highlight both the best and second-best performing combinations to facilitate comparative analysis. Notably, no single combination consistently outperforms others across all test datasets. However, nearly all combinations attain either the best or second-best results on at least one test set, suggesting that each pairing contributes complementary information under different distributional shifts. These observations indicate that combinations encompassing more diverse domains tend to provide more transferable cues across heterogeneous OOD scenarios.

\begin{table*}[ht]
\centering
\small  
\caption{OOD detection performance using pairwise combinations of \textit{atypical} data subsets, each introducing varying degrees of semantic and visual coverage. The best results per metric are highlighted in \textbf{bold}, and the second-best results are \underline{underlined}. The symbol ``+'' indicates the introduction of auxiliary \textit{atypical} data. Abbreviations: abn = abnormal, sci = sci-fi, uni = unintentional, ani = animation.}

\vspace{0.5em}
\resizebox{\textwidth}{!}{
\begin{tabularx}{\textwidth}{c|c|*{6}{>{\centering\arraybackslash}X}|c}

\toprule
\footnotesize \textbf{Test Sets} & 
\footnotesize \textbf{Metric} & 
\footnotesize +abn\_sci & 
\footnotesize +abn\_uni & 
\footnotesize +abn\_ani & 
\footnotesize +ani\_sci & 
\footnotesize +ani\_uni & 
\footnotesize +sci\_uni & 
\footnotesize \makecell{Mean ± Std Dev} \\
\midrule
\multirow{3}{*}
{\shortstack{{HMDB51}\\~\cite{kuehne2011hmdb}}} 
& FPR95 ↓ & 81.86 & \underline{78.36} & 80.72 & 82.06 & 79.06 & \textbf{76.77} & $79.81\pm2.10$ \\
& AUROC ↑ & 65.50 & 65.77 & 66.63 & 66.03 & \underline{68.09} & \textbf{69.97} & $67.00\pm1.72$ \\
& AUPR ↑  & \underline{29.38} & 23.39 & 28.40 & \textbf{30.49} & 25.93 & 28.67 & $27.71\pm2.60$ \\
\hline
\multirow{3}{*}{
\shortstack{{MiT-v2}\\~\cite{monfort2019moments}}} 
& FPR95 ↓ & 79.98 & 68.26 & 78.49 & 79.70 & \textbf{61.40} & \underline{64.83} & $72.11\pm8.27$ \\
& AUROC ↑ & 63.89 & 73.46 & 67.23 & 65.35 & \textbf{75.30} & \underline{74.87} & $70.02\pm5.11$ \\
& AUPR ↑  & 24.36 & 31.82 & 26.91 & 25.36 & \underline{32.94} & \textbf{33.16} & $29.09\pm4.00$ \\
\hline
\multirow{3}{*}{\shortstack{{\textnormal{[}\textit{atypical}\textnormal{]}} \\-surreal}} 
& FPR95 ↓ & 75.99 & 48.32 & 73.31 & 74.11 & \textbf{47.32} & \underline{48.29} & $61.22\pm14.54$ \\
& AUROC ↑ & 64.54 & \textbf{83.28} & 67.74 & 65.95 & \underline{82.94} & 82.22 & $74.45\pm9.23$ \\
& AUPR ↑  & 22.26 & \textbf{44.81} & 25.18 & 23.31 & \underline{44.69} & 44.14 & $34.07\pm11.52$ \\
\hline
\multirow{3}{*}{\shortstack{{\textnormal{[}\textit{atypical}\textnormal{]}} \\-theatre}} 
& FPR95 ↓ & 75.96 & 81.27 & \underline{57.40} & \textbf{41.13} & 58.72 & 64.70 & $63.20\pm14.37$ \\
& AUROC ↑ & 69.89 & 56.26 & \underline{82.45} & \textbf{91.18} & 73.34 & 70.19 & $73.89\pm11.94$ \\
& AUPR ↑  & 31.75 & 16.73 & \underline{52.84} & \textbf{73.82} & 28.64 & 26.72 & $38.42\pm21.02$ \\
\bottomrule
\end{tabularx}
}
\label{tab:ood2}
\vspace{-1em}
\end{table*}

\section{Conclusion}
In this paper, for the first time, we investigated the potential of atypical videos for representation learning in open-world scenarios, and found that they offer valuable insights and improvements for open-world understanding. To address the limitations of existing datasets in representing non-conventional video content, we introduce a novel dataset, termed \textit{atypical}, which comprises a diverse collection of videos that deviate from standard, well-defined categories. This dataset was introduced aiming to better address the challenges of open-world scenarios and to explore its impact on the critical task of OOD detection, NCD and ZSAR. Our experiments suggested that training with a smaller, yet diverse set of atypical samples substantially improves the robustness of the model in identifying novel data distributions. The diversity within the atypical dataset played a crucial role in driving these improvements, underscoring the importance of extending traditional datasets with more varied and unconventional content. Looking ahead, atypical data presents several promising avenues for future research. One potential direction is the continued enrichment of these datasets to better capture the unpredictability of real-world environments. Furthermore, developing adaptive learning techniques that integrate new atypical samples during inference could enable models to evolve dynamically, maintaining resilience in ever-changing conditions. 
We hope the newly introduced atypical dataset in this paper will facilitate new directions in advancing open-world understanding and learning.

\section*{Acknowledgements}

This project is partially supported by the Royal Society grants (SIF\textbackslash R1\textbackslash231009, IES\textbackslash R3\textbackslash223050) and an Amazon Research Award.
The computations in this research were performed using the Baskerville Tier 2 HPC service. Baskerville was funded by the EPSRC and UKRI through the World Class Labs scheme (EP\textbackslash T022221\textbackslash1) and the Digital Research Infrastructure programme (EP\textbackslash W032244\textbackslash1) and is operated by Advanced Research Computing at the University of Birmingham. 


\newpage
\appendix
\renewcommand{\thefigure}{\thesection.\arabic{figure}}
\renewcommand{\thetable}{\thesection.\arabic{table}}


\vspace{3em}
\begin{flushleft}
    {\LARGE \sffamily
    \textcolor{customblue}
    {\textbf{\textit{What Can We Learn from Harry Potter?}}}\\[4pt]
    \textcolor{customblue}{\textbf{An Exploratory Study of}}\\[2pt]
    \textcolor{customblue}{\textbf{Visual Representation Learning from}}\\[4pt]
    \textcolor{customblue}{\textbf{Atypical Videos}}}
\end{flushleft}

\begin{center}
    {\Large \sffamily \textcolor{customblue}{\textbf{Supplementary Material}}} 
\end{center}

\noindent
\noindent
\begin{minipage}[t]{0.48\textwidth}
    \sffamily
    \raggedright
    Qiyue Sun\textsuperscript{\scriptsize 1,2}\\
    {\small \textcolor{customblue}{julysun@mail.sdu.edu.cn}}\\[0.5em]
    Qiming Huang\textsuperscript{\scriptsize 1}\\
    {\small \textcolor{customblue}{qxh366@student.bham.ac.uk}}\\[0.5em]
    Yang Yang\textsuperscript{\scriptsize 2}\\
    {\small \textcolor{customblue}{yyang@sdu.edu.cn}}\\[0.5em]
    Hongjun Wang\textsuperscript{\scriptsize 2}\\
    {\small \textcolor{customblue}{hjw@sdu.edu.cn}}\\[0.5em]
    Jianbo Jiao\textsuperscript{\scriptsize 1}\\
    {\small \textcolor{customblue}{j.jiao@bham.ac.uk}}
\end{minipage}%
\hfill
\begin{minipage}[t]{0.48\textwidth}
    \sffamily
    \raggedright
    \begin{tabular}[t]{@{}l@{\hspace{1pt}}l@{}}
        \textsuperscript{\scriptsize 1} & The MIx Group, School of Computer\\
                                        & Science, University of Birmingham, \\ 
                                        & Birmingham, UK \\[1em]
        \textsuperscript{\scriptsize 2} & School of Information Science and\\ 
                                        & Engineering, Shandong University, \\
                                        & Qingdao, China
    \end{tabular}
\end{minipage}

\section{\textit{Atypical}}
\label{Appendix_A}
\setcounter{figure}{0}
\setcounter{table}{0}
\RaggedRight
Here, we present examples of \textit{atypical} data to highlight their diversity. Figure~\ref{fig:sci-fi} shows sci-fi videos with futuristic visuals, while Figure~\ref{fig:animation} shows a variety of animations ranging from photorealistic to stylised designs. Figure~\ref{fig:unintentional} illustrates unintentional events, and Figure~\ref{fig:abnormal} depicts anomalies like criminal activities that deviate from typical datasets. Figure~\ref{fig:surreal} shows synthetic data with virtual characters, and Figure~\ref{fig:theatre} highlights theatrical performances with exaggerated designs and actions. \rev{More details can be found in our project page: \url{https://julysun98.github.io/atypical_dataset/}}.

\begin{center}
    \centering
    \includegraphics[width=0.9\textwidth]{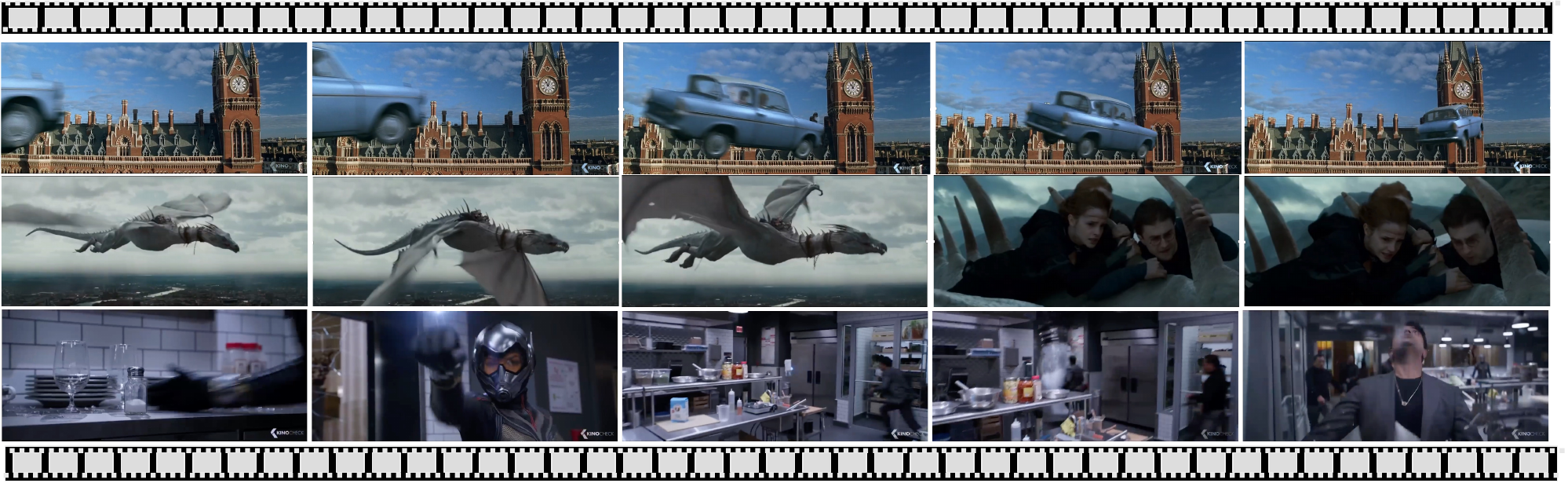}
    \vspace{0.5em}
    \captionof{figure}{[\textit{atypical}]-sci-fi}
    \label{fig:sci-fi}
\end{center}

\begin{center}
    \centering
    \includegraphics[width=0.9\textwidth]{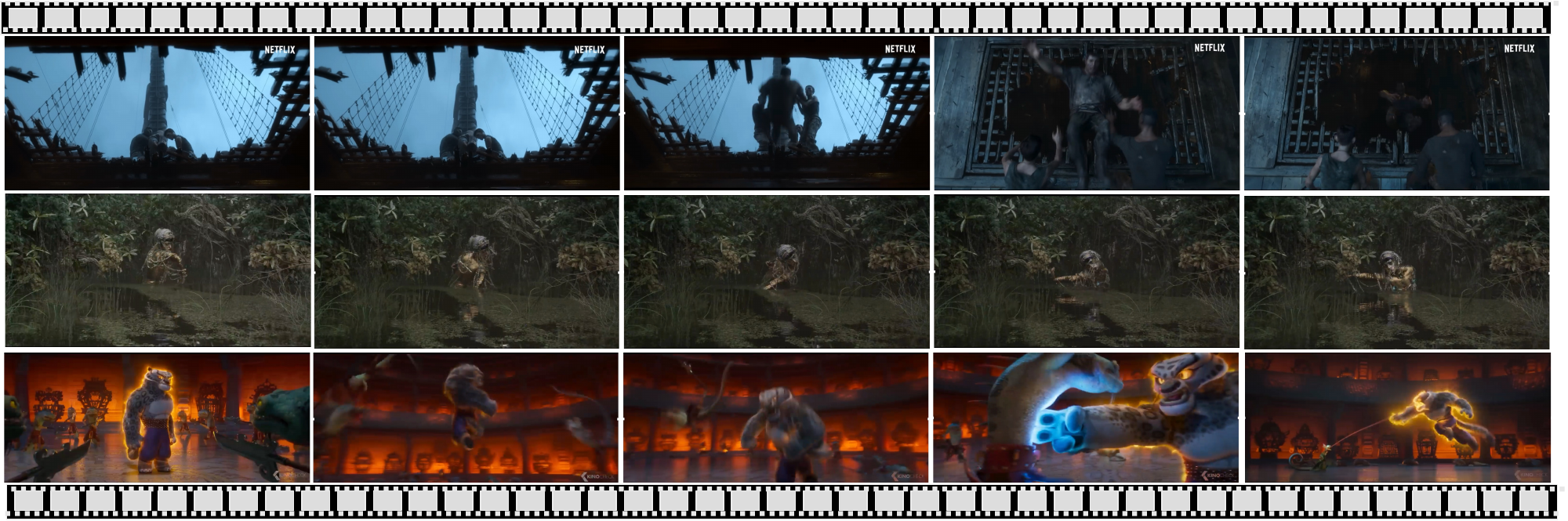}
    \vspace{0.5em}
    \captionof{figure}{[\textit{atypical}]-animation}
    \label{fig:animation}
\end{center}
\begin{center}
    \centering
    \includegraphics[width=0.9\textwidth]{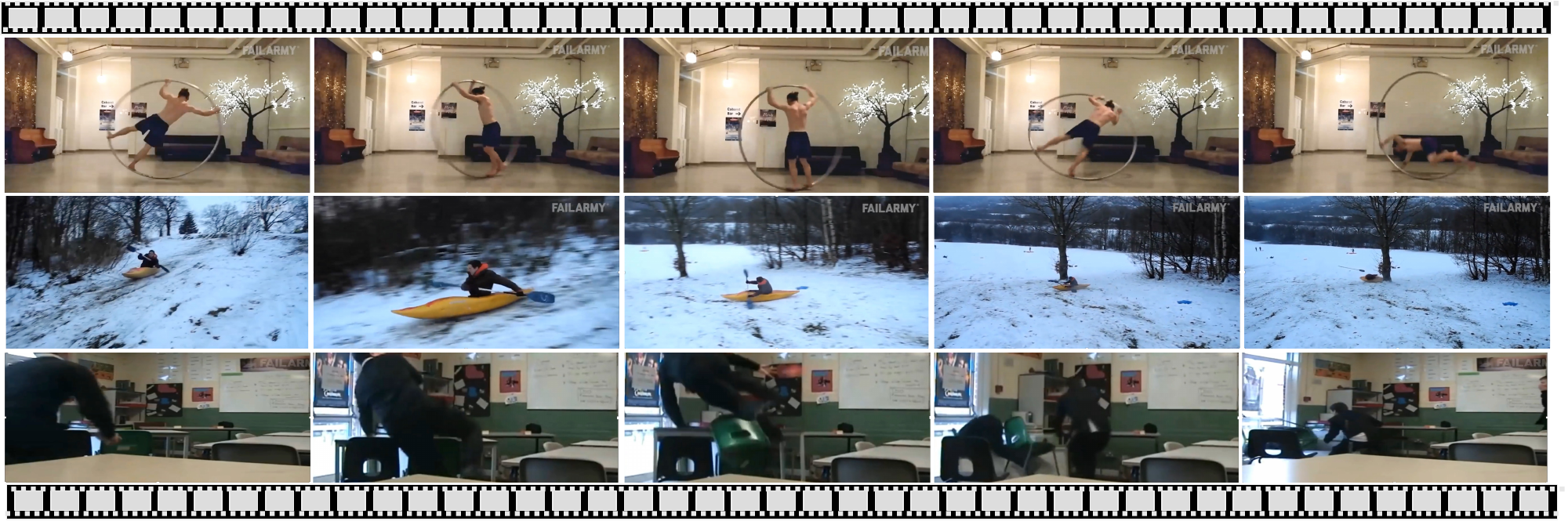}
    \vspace{0.5em}
    \captionof{figure}{[\textit{atypical}]-unintentional}
    \label{fig:unintentional}
\end{center}

\begin{center}
    \centering
    \includegraphics[width=0.9\textwidth]{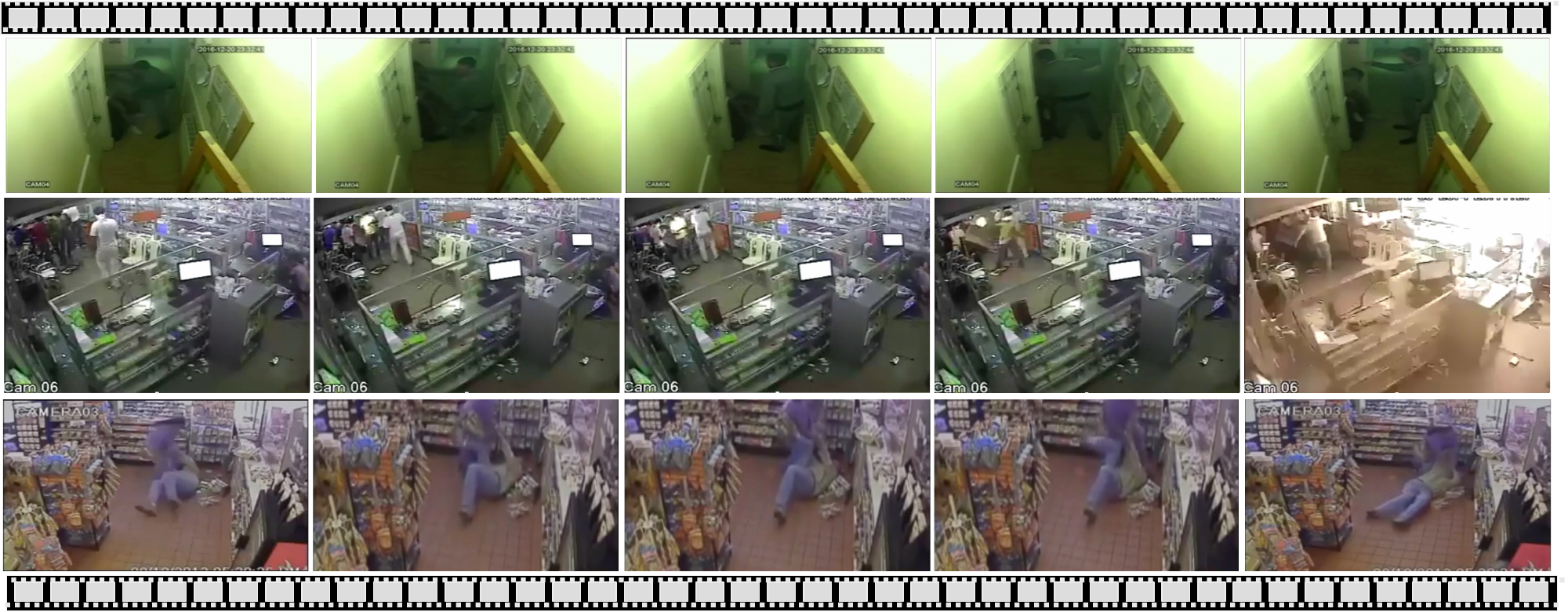}
    \vspace{0.5em}
    \captionof{figure}{[\textit{atypical}]-abnormal}
    \label{fig:abnormal}
\end{center}

\begin{center}
    \centering
    \includegraphics[width=0.9\textwidth]{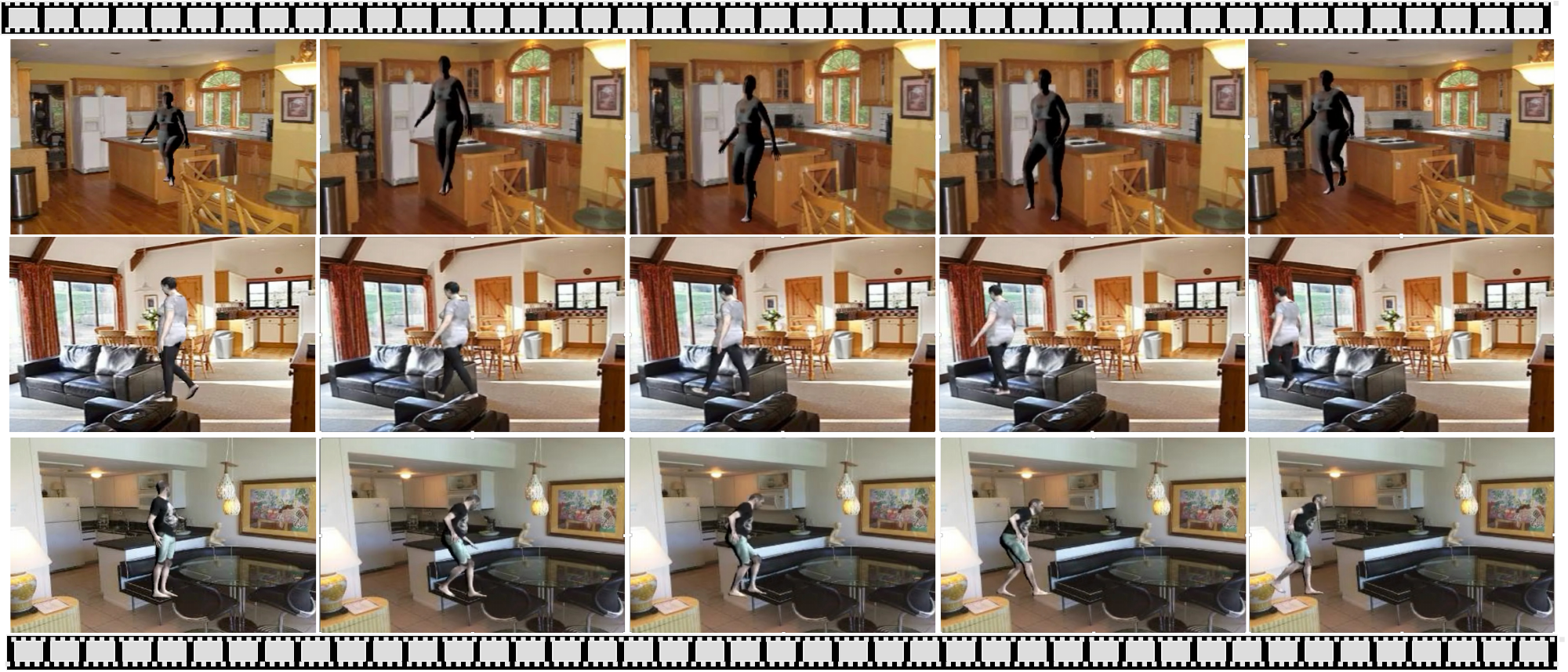}
    \vspace{0.5em}
    \captionof{figure}{[\textit{atypical}]-surreal}
    \label{fig:surreal}
\end{center}

\begin{center}
    \centering
    \includegraphics[width=0.9\textwidth]{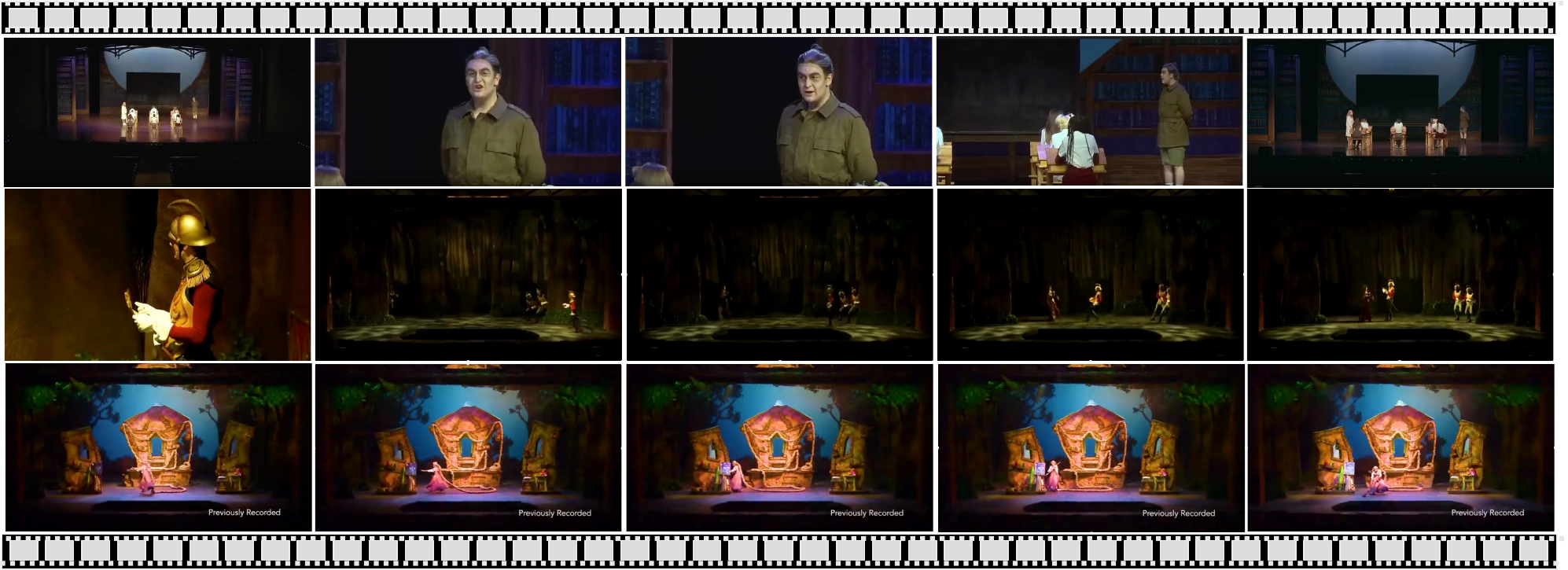}
    \vspace{1em}
    \captionof{figure}{[\textit{atypical}]-theatre}
    \label{fig:theatre}
\end{center}
\vspace{-1em}
\section{Preliminaries}
\label{Appendix_tasks}
\setcounter{figure}{0}
\setcounter{table}{0}
\subsection{Out-of-distribution (OOD) detection}

Out-of-distribution (OOD) detection can be formulated as a binary classification problem, where the goal is to determine whether a test sample \( x \in X \) originates from the in-distribution dataset \( D_{\text{in}} \) (ID) or from an unknown distribution (OOD)~\citep{hendrycks2016baseline}. A sample is predicted as OOD if its confidence score \( P(x \mid \text{ID}) \) is lower than a threshold \( \tau \), as defined below:

\[
\text{OOD}(x) =
\begin{cases}
1, & \text{if } P(x \mid \text{ID}) < \tau \\
0, & \text{otherwise}
\end{cases}
\]

Outlier exposure (OE)~\citep{hendrycks2018deep,ming2022poem,zhu2023diversified} is a common approach that introduces auxiliary outlier data \( D_{\text{aux}} \) to help the model learn to distinguish OOD samples. A \( k \)-class classifier \( f \) is trained using labelled in-distribution data, while OE further incorporates unlabeled auxiliary data by encouraging the model to produce uniform predictions on OOD inputs. The objective is to minimise the expected cross-entropy between the model output \( f(x) \) and a uniform distribution \( \mathcal{U} \):

\[
\mathbb{E}_{x \sim \mathcal{D}_{\text{aux}}} \left[ H(\mathcal{U}; f(x)) \right]
\]

where \( H \) denotes cross-entropy.
\vspace{-1em}
\subsection{Novel category discovery (NCD)}

Novel Category Discovery (NCD) aims to identify and cluster previously unseen categories from an unlabelled dataset. Given a labelled dataset $D_l = \{(x_l^i, y_l^i)\}$ containing known classes, and an unlabelled dataset $D_u = \{x_u^i\}$ with novel classes, where the class sets are disjoint, the goal is to cluster the instances in \( D_u \) without access to their ground-truth labels.

AutoNovel~\citep{han2021autonovel} introduces a model \( \Phi \) trained on \( D_l \) to obtain generalizable representations for use in clustering \( D_u \), under the assumption that knowledge learned from known categories transfers to unknown ones. The process consists of three steps: 1) \( \Phi \) is pretrained using a self-supervised learning objective (e.g., RotNet~\citep{gidaris2018unsupervised}) on \( D_l \cup D_u \); 2) The pretrained encoder is extended with a classification head $\eta^l: \mathbb{R}^d \rightarrow \mathbb{R}^{C_l}$ and fine-tuned on \( D_l \) using cross-entropy loss with ground-truth labels \( y_l^i \); 3) Pseudo labels are generated for \( D_u \) based on top-\( k \) ranking agreement, where the similarity between two samples is defined as:
\[
s^{ij} = \mathbb{I} \left\{ \text{top}_k(\Phi(x_u^i)) = \text{top}_k(\Phi(x_u^j)) \right\};
\]
3) The model \( \Phi \) is then jointly trained on \( D_l \) with ground-truth labels and on \( D_u \) with pseudo labels \( s^{ij} \) for clustering.
\vspace{-1em}
\subsection{Zero-shot action recognition (ZSAR)}

Zero-Shot Action Recognition (ZSAR) aims to recognise unseen action categories without using labelled examples from those classes. Given a labeled dataset $D_s = \{(v_s^i, y_s^i)\}$ with seen classes \( C_s \), and an unlabeled dataset $D_t = \{v_t^i\}$ containing unseen classes \( C_t \), where \( C_s \cap C_t = \emptyset \), the objective is to train a model that generalizes to \( C_t \) based on semantic alignment.

The process involves two main stages: 1) A visual encoder \( \Phi \) is pretrained on \( D_s \cup D_t \) using self-supervised contrastive learning, where two augmented views of each video are encoded and pulled closer in the feature space while being pushed apart from other videos. The objective is to maximise $\cos(\Phi(v_i), \Phi(v_i^+))$ over positive pairs, while minimising it over negatives, scaled by a temperature \( \tau \). 2) The encoder is fine-tuned on \( D_s \) by aligning video embeddings with semantic class embeddings \( \psi(c) \) derived from multiple natural language prompts. Each class name is mapped to text embeddings via a frozen text encoder and averaged across prompts. Training minimises a softmax-based classification loss using cosine similarity:
\[
\mathcal{L}_{\text{align}} = - \log \frac{\exp(\cos(\Phi(v_s^i), \psi(y_s^i)) / \tau)}{\sum_{c \in C_s} \exp(\cos(\Phi(v_s^i), \psi(c)) / \tau)}.
\]

At test time, a video \( v_t \) is classified by assigning it the unseen class with the highest similarity score:
\[
\hat{y}_t = \arg\max_{c \in C_t} \cos(\Phi(v_t), \psi(c)).
\]

\vspace{-1em}
\section{Datasets and experiments}
\setcounter{figure}{0}
\setcounter{table}{0}
\subsection{Detail about datasets splits}
\label{appendix_datasets_splits}
For OOD detection task: To ensure a clear distinction between ID and OOD categories, we followed the method proposed by ~\citep{hendrycks2018deep,cen2023enlarging} to remove the overlap categories between datasets. Specifically, we removed 6 overlapping action categories in HMDB51~\citep{kuehne2011hmdb} and UCF101, as well as 93 overlapping actions between Kinetics-400~\citep{kay2017kinetics} and UCF101~\citep{soomro2012ucf101} and HMDB51~\citep{kuehne2011hmdb}. In addition, 33 categories from the MiT-v2 dataset that were not present in the other three datasets were selected for testing as OOD data. The categories below were removed before training.

\textbf{HMDB51~\citep{kuehne2011hmdb}}: 35, Shoot bow; 29, Push up; 15, Golf; 26, Pull up; 30, Ride bike; 34, Shoot ball; 43, Swing baseball; 31, Ride horse.

\textbf{UCF101~\citep{soomro2012ucf101}}: 2, Archery; 71, PushUps; 32, GolfSwing; 69, PullUps; 10, Biking; 7, Basketball; 6, BaseballPitch; 41, HorseRiding.

\textbf{Kinetics-400~\citep{kay2017kinetics}}: 3, applauding; 5, arm wrestling; 18, auctioning; 19, baby waking up; 22, balloon blowing; 27, beatboxing; 31, bending back; 36, biking through snow; 40, blowing leaves; 45, bookbinding; 48, bouncing on trampoline; 49, bowling; 57, brushing teeth; 66, carrying baby; 67, cartwheeling; 68, catching or throwing baseball; 77, catching or throwing frisbee; 91, catching or throwing softball; 93, celebrating; 99, changing oil; 100, changing wheel; 101, checking tires; 102, cheerleading; 107, chopping wood; 108, clapping; 109, clay pottery making; 110, clean and jerk; 111, cleaning floor; 112, cleaning gutters; 113, cleaning pool; 114, cleaning shoes; 115, cleaning toilet; 116, cleaning windows; 117, climbing a rope; 138, climbing tree; 141, cooking chicken; 142, cooking egg; 143, cooking on campfire; 147, counting money; 148, country line dancing; 151, cracking neck; 153, crawling baby; 154, crossing river; 158, cutting pineapple; 159, cutting watermelon; 166, dancing ballet; 169, dancing gangnam style; 171, deadlifting; 174, decorating the christmas tree; 175, digging; 176, dining; 179, disc golfing; 180, diving cliff; 182, dodgeball; 188, dribbling basketball; 220, dunking basketball; 221, dying hair; 223, eating cake; 227, eating ice cream; 230, egg hunting; 231, exercising arm; 232, exercising with an exercise ball; 237, feeding fish; 241, filling eyebrows; 246, fixing hair; 250, folding clothes; 251, folding napkins; 255, front raises; 258, gargling; 259, getting a haircut; 260, getting a tattoo; 273, giving or receiving award; 278, golf chipping; 296, grooming horse; 297, gymnastics tumbling; 305, hammer throw; 306, headbanging; 307, headbutting; 308, high jump; 309, high kick; 310, hitting baseball; 311, hockey stop; 312, holding snake; 322, hugging; 323, hula hooping; 325, ice climbing; 329, ice skating; 330, ironing; 339, javelin throw; 340, jetskiing; 345, juggling balls; 357, kissing; 367, laying bricks; 378, long jump; 395, making a sandwich; 396, writing.

And the categories of \textbf{MiT-v2~\citep{monfort2019moments}} below were selected: 2, burying; 3, covering; 4, flooding; 12, submerging; 13, breaking; 16, destroying; 17, competing; 18, giggling; 21, flicking; 34, locking; 37, flipping; 38, sewing; 39, clipping; 47, constructing; 50, screwing; 51, shrugging; 53, cracking; 54, scratching; 56, selling; 60, clinging; 87, bubbling; 88, joining; 97, kneeling; 151, peeling; 153, wetting; 159, inflating; 168, launching; 172, leaking; 205, overflowing; 221, storming; 255, combusting; 296, cramming; 297, burning.


\vspace{-2em}
\subsection{Implementation details.}
\label{Appendix_Implementation}
For the OOD detection task, we use the outlier exposure method~\citep{hendrycks2018deep} to introduce auxiliary data during training. The baseline is trained using only ID data with a cross-entropy loss for multi-class classification over 100 epochs. In the outlier exposure setting, we fine-tune the pre-trained baseline model by introducing various auxiliary datasets. The initial learning rate is set to 0.1 and decays following a cosine learning rate schedule.
For the NCD task, we follow the method of AutoNovel~\citep{han2021autonovel}. In (1) the self-supervised pre-training step, we use the pace prediction task~\citep{wang2020self} to pre-train video representations for 200 epochs. At (2) the fine-tuning step with labelled data, we fine-tune for 100 epochs. Finally, at (3) the joint training step with labelled and unlabeled data, we fine-tune our model for 200/100 epochs on UCF101~\citep{soomro2012ucf101} and HMDB51~\citep{kuehne2011hmdb}, respectively. The initial learning rate for all experiments is set to 0.01 and halved every 10 epochs.
For the ZSAR task, we adopt a three-stage training pipeline. In the (1) self-supervised pre-training stage, we train a temporal video encoder using contrastive learning on unlabeled video data for 200 epochs, keeping the CLIP~\citep{radford2021learning} image encoder frozen. In the (2) supervised training stage, we fine-tune the temporal module on labelled base classes using KL divergence loss to align video features with text-based prompts. Finally, in the (3) zero-shot evaluation stage, we use the learned video-text space to recognise unseen classes based solely on their semantic names without any fine-tuning.
\vspace{-3em}
\subsection{Experiment results}
\subsubsection{OOD detection}
\label{Appendix_OOD}
We replicate the same experiments using a ViT-based TimeSformer backbone~\citep{gberta_2021_ICML}, with results reported in Table \ref{tab:OOD_main_timesformer}. The experimental results are consistent with those obtained using ResNet3D-50~\citep{hara2018can}, further demonstrating the effectiveness of our atypical data in enhancing performance on out-of-distribution detection tasks.
\vspace{-1em}

\begin{table}[H]
\centering
\small
\caption{OOD detection performance using different auxiliary datasets with a ViT-based TimeSformer backbone. Results are reported on four test sets: HMDB51, MiT-v2, [\textit{atypical}]-surreal, and [\textit{atypical}]-theatre. \textit{Baseline}: training w/o auxiliary data; the others add auxiliary data. The best results are highlighted in \textbf{bold}.}
\vspace{1em}
\setlength{\tabcolsep}{5pt}
\begin{tabular}{ll|ccccc}
\toprule
\textbf{Dataset} & \textbf{Metric} & Baseline & +Gaussian & +Diving48 & +K400 & +\textit{Atypical} \\
\midrule
\multirow{3}{*}{HMDB51~\cite{kuehne2011hmdb}}
& FPR95 ↓ & 87.09 & 88.57 & 89.14 & 87.14 & \textbf{85.39} \\
& AUROC ↑ & 57.36 & 57.80 & 55.66 & 60.96 & \textbf{61.63} \\
& AUPR ↑  & 19.88 & 20.45 & 18.71 & 22.81 & \textbf{24.67} \\
\midrule
\multirow{3}{*}{MiT-v2~\cite{monfort2019moments}}
& FPR95 ↓ & 89.11 & 89.00 & 88.45 & 86.28 & \textbf{85.39} \\
& AUROC ↑ & 55.26 & 56.05 & 57.09 & 60.97 & \textbf{61.63} \\
& AUPR ↑  & 18.81 & 19.37 & 20.71 & 23.27 & \textbf{24.67} \\
\midrule
\multirow{3}{*}{[\textit{atypical}]-surreal}
& FPR95 ↓ & 91.18 & 91.78 & 93.28 & 91.19 & \textbf{85.16} \\
& AUROC ↑ & 53.11 & 53.20 & 49.65 & 50.85 & \textbf{61.78} \\
& AUPR ↑  & 17.86 & 17.97 & 15.89 & 15.99 & \textbf{25.24} \\
\midrule
\multirow{3}{*}{[\textit{atypical}]-theatre}
& FPR95 ↓ & 88.44 & 88.64 & 86.34 & 81.28 & \textbf{79.41} \\
& AUROC ↑ & 56.44 & 56.54 & 55.95 & 68.30 & \textbf{71.11} \\
& AUPR ↑  & 19.28 & 19.64 & 18.12 & 29.14 & \textbf{34.62} \\
\bottomrule
\end{tabular}
\vspace{-3em}
\label{tab:OOD_main_timesformer}
\end{table}

\vspace{1em}
\subsubsection{Novel category discovery}
\label{Appendix_NCD}
Evaluation on the HMDB51~\citep{kuehne2011hmdb} dataset is presented in Table~\ref{tab:ND_main_HMDB}. The results are consistent with those observed on UCF101~\citep{soomro2012ucf101}, further confirming the effectiveness of the proposed \textit{atypical} dataset in improving novel category discovery.


\vspace{-0.5em}
\begin{table}[h]
\centering
\caption{NCD performance on HMDB51 with different auxiliary datasets used for self-supervised pre-training. \textit{Baseline}: training w/o auxiliary data. HMDB51 refers to using the original dataset itself as auxiliary data. The best results are highlighted in \textbf{bold}.}
\vspace{1em}
\small
\resizebox{\textwidth}{!}{%
\begin{tabular}{l|c|ccc|cc}
\toprule
\textbf{Metric} & Baseline &+HMDB51 & +K400 &   +\textit{atypical} & +HMDB51+K400 & +HMDB51+\textit{atypical} \\
\midrule
acc ↑  & 18.18  & 22.12 & 20.00 & 23.03 & 23.03 & \textbf{26.67} \\
nmi ↑  & 0.1632 & 0.1311 & 0.1080 & 0.1525 & 0.1821 & \textbf{0.1922} \\
ari ↑  &0.0317 & 0.0495 & 0.0341 &  0.0450 & \textbf{0.0720} & 0.0692 \\
\bottomrule
\end{tabular}
}
\vspace{-0.5em}
\label{tab:ND_main_HMDB}
\end{table}

\begin{figure*}[ht]
    \begin{center} \includegraphics[width=.97\textwidth, trim={0  0 0  2em}, clip]{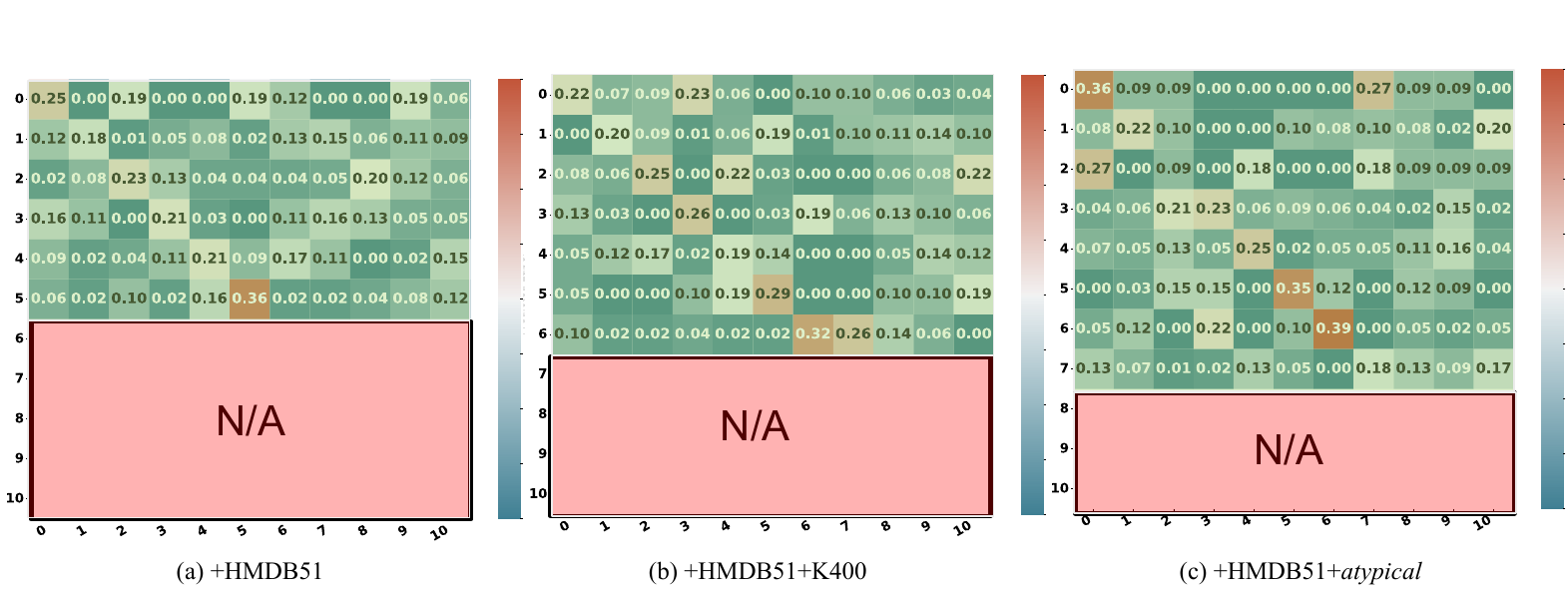}
    \end{center}
    \caption{Confusion matrix on unlabelled classes of HMDB51. The horizontal axis represents the True label and the vertical axis represents the predicted label. + denotes pretraining with different datasets. N/A denotes that there is no predicted output for the class. Green cells highlight novel classes that the model discovered, while red cells denote novel classes that were not learned.}
    \label{fig:ND_ablation}
    \vspace{-1em}
\end{figure*}

To evaluate the effect of auxiliary data on representation quality, we conduct an NCD experiment under different self-supervised pre-training settings. Specifically, we compare three configurations: (1) using HMDB51~\citep{kuehne2011hmdb} alone, (2) HMDB51~\citep{kuehne2011hmdb} combined with Kinetics-400~\citep{kay2017kinetics}, and (3) HMDB51~\citep{kuehne2011hmdb} combined with our proposed \textit{atypical} dataset. All settings use the same labelled-unlabelled splits and training protocol. As shown in Figure~\ref{fig:ND_ablation}, the model pre-trained with atypical data shows reduced class confusion—evident from lower off-diagonal values—and more balanced utilisation of all clusters compared to HMDB51~\citep{kuehne2011hmdb} and K400, as reflected by the denser and more uniformly distributed non-zero entries across predicted labels.

\vspace{-1em}

\begin{table}[htbp]
\centering
\caption{ZSAR performance on the UCF\_21 split with different auxiliary datasets used for representation learning. \textit{Baseline}: training w/o auxiliary data. The best results are highlighted in \textbf{bold}.}
\vspace{1em}
\small
\resizebox{\textwidth}{!}{%
\begin{tabular}{l|c|ccc|cc}
\toprule
\textbf{Metric} & Baseline & +UCF101 & +K400   & +\textit{atypical} & +UCF101+K400 & +UCF101+\textit{atypical} \\
\midrule
Top-1 ↑  & 51.01  & 53.68 & 54.78   & 56.9   & 57.20 & \textbf{57.56} \\
Top-5 ↑  & 89.05 & 91.80 & 95.24  &  93.26 & \textbf{95.46} & 94.10 \\
\bottomrule
\end{tabular}
}
\vspace{-2em}
\label{tab:ZSAR_main_ucf}
\end{table}

\subsubsection{Zero-shot action recognition}
\label{Appendix_ZSAR}
Similarly, Table~\ref{tab:ZSAR_main_ucf} presents the results of ZSAR on UCF101~\citep{soomro2012ucf101}. Consistent with previous findings on HMDB51~\citep{kuehne2011hmdb}, the combination of UCF101~\citep{soomro2012ucf101} and the proposed \textit{atypical} dataset yields the highest performance.

\subsubsection{Semantic diversity helps open-world learning}
\label{Appendix_diversity}
The detailed results corresponding to Figure 5 in the main paper are presented in Tables~\ref{tab:ood1} and \ref{tab:ood3}. $abn$ denotes [\textit{atypical}]-{abnormal}, $ani$ denotes [\textit{atypical}]-animation, $sci$ denotes [\textit{atypical}]-sci-fi, and $uni$ denotes [\textit{atypical}]-unintentional. We introduced different categories of atypical data to investigate the impact of various data categories on improving OOD detection performance.

In Table~\ref{tab:ood1}, we introduced only one type of atypical data. From the results, we observed that using a single category of atypical data may improve OOD detection performance for a specific $D_{out}$, but it fails to enhance detection performance across all $D_{out}$ distributions. For example, introducing [\textit{atypical}]-sci-fi data significantly improved OOD detection performance on HMDB51~\citep{kuehne2011hmdb} but was ineffective for OOD detection on MiT-v2~\citep{monfort2019moments}. Furthermore, as the variety of introduced atypical data increases (as shown in Table 6 in the main paper and Table~\ref{tab:ood3}), the OOD detection performance across different $D_{out}$ improves significantly, and the standard deviation decreases correspondingly. This indicates that incorporating more diverse atypical data not only enhances OOD detection performance across various $D_{out}$ distributions but also leads to more stable detection performance for the same $D_{out}$.
\begin{table}[htbp]
\centering
\small
\caption{OOD detection performance with the introduction of any one category of \textit{atypical} data. + indicates the introduction of auxiliary \textit{atypical} data.  $abn$ denotes [\textit{atypical}]-{abnormal}, $ani$ denotes [\textit{atypical}]-animation, $sci$ denotes [\textit{atypical}]-sci-fi, and $uni$ denotes [\textit{atypical}]-unintentional. The best results per metric are highlighted in \textbf{bold}, and the second-best results are \underline{underlined}.}
\vspace{1em}
\begin{tabular}{c|c|cccc|c}
\hline
\makecell{Test sets} & Metric & +$abn$ &  +$ani$  & +$sci$ & +$uni$ & \makecell{Mean ± Std Dev}\\
\hline
\multirow{3}{*}{\shortstack{{HMDB51}\\~\cite{kuehne2011hmdb}}}  
& FPR95 ↓ & 84.41 & \underline{80.41} & \textbf{79.06} & 81.29 & $81.29\pm2.27$ \\
& AUROC ↑ & 60.81 & \underline{67.01} & \textbf{68.14} & 66.05 & $65.50\pm3.24$ \\
& AUPR ↑  & 21.73 & \underline{28.09} & \textbf{32.36} & 23.84 & $26.50\pm4.71$ \\
\hline
\multirow{3}{*}{
\shortstack{{MiT-v2}\\~\cite{monfort2019moments}}}
& FPR95 ↓ & 83.63 & \underline{75.75} & 81.93 & \textbf{61.05} & $75.59\pm10.27$ \\
& AUROC ↑ & 61.98 & \underline{66.86} & 62.85 & \textbf{74.39} & $66.52\pm5.66$ \\
& AUPR ↑  & 22.98 & \underline{25.42} & 23.45 & \textbf{31.48} & $25.83\pm3.91$ \\
\hline
\multirow{3}{*}{\shortstack{{\textnormal{[}\textit{atypical}\textnormal{]}} \\-surreal}} 
& FPR95 ↓ & 75.52 & \underline{69.46} & 75.43 & \textbf{41.80} & $65.55\pm16.09$ \\
& AUROC ↑ & 66.78 & \underline{69.86} & 66.41 & \textbf{84.93} & $72.00\pm8.76$ \\
& AUPR ↑  & 24.34 & \underline{25.85} & 24.25 & \textbf{47.11} & $30.39\pm11.17$ \\
\hline
\multirow{3}{*}{\shortstack{{\textnormal{[}\textit{atypical}\textnormal{]}} \\-theatre}} 
& FPR95 ↓ & 88.53 & \textbf{40.78} & \underline{67.05} & 70.75 & $66.78\pm19.71$ \\
& AUROC ↑ & 53.71 & \textbf{89.66} & \underline{78.73} & 59.29 & $70.35\pm16.76$ \\
& AUPR ↑  & 16.60 & \textbf{64.97} & \underline{48.42} & 17.45 & $36.86\pm23.88$ \\
\hline
\end{tabular}
\vspace{-2em}
\label{tab:ood1}
\end{table}
\begin{table}[H]
\centering
\small
\caption{OOD detection performance with the introduction of any three categories of \textit{atypical} data. + indicates the introduction of auxiliary \textit{atypical} data.  $abn$ denotes [\textit{atypical}]-{abnormal}, $ani$ denotes [\textit{atypical}]-animation, $sci$ denotes [\textit{atypical}]-sci-fi, and $uni$ denotes [\textit{atypical}]-unintentional. The best results per metric are highlighted in \textbf{bold}, and the second-best results are \underline{underlined}.}
\vspace{0.5em}
\resizebox{\textwidth}{!}{
\begin{tabular}{c|c|cccc|c}
\hline
\makecell{Test sets} & Metric & +$ani\_abn\_sci$ & +$ani\_abn\_uni$ & +$ani\_sci\_uni$ & +$abn\_sci\_uni$ & \makecell{Mean ± Std Dev} \\
\hline
\multirow{3}{*}{\shortstack{{HMDB51}\\~\cite{kuehne2011hmdb}}}  
& FPR95 ↓ & 80.53 & \textbf{73.21} & \underline{75.68} & 76.51 & $76.48 \pm 3.04$ \\
& AUROC ↑ & 66.84 & 67.19 & \textbf{69.44} & \underline{67.77} & $67.81 \pm 1.15$ \\
& AUPR ↑  & \textbf{29.81} & 24.49 & \underline{27.79} & 25.61 & $26.93 \pm 2.36$ \\
\hline
\multirow{3}{*}{
\shortstack{{MiT-v2}\\~\cite{monfort2019moments}}} 
& FPR95 ↓ & 77.76 & 68.28 & \textbf{65.06} & \underline{66.24} & $69.34 \pm 5.77$ \\
& AUROC ↑ & 67.12 & 72.64 & \underline{73.89} & \textbf{73.98} & $71.91 \pm 3.25$ \\
& AUPR ↑  & 25.85 & 30.40 & \underline{31.59} & \textbf{32.35} & $30.05 \pm 2.91$ \\
\hline
\multirow{3}{*}{\shortstack{{\textnormal{[}\textit{atypical}\textnormal{]}} \\-surreal}} 
& FPR95 ↓ & 71.84 & \underline{50.31} & \textbf{48.93} & 52.17 & $55.81 \pm 10.77$ \\
& AUROC ↑ & 66.89 & 81.35 & \textbf{83.42} & \underline{81.44} & $78.28 \pm 7.65$ \\
& AUPR ↑  & 23.63 & 41.54 & \textbf{46.17} & \underline{42.03} & $38.34 \pm 10.03$ \\
\hline
\multirow{3}{*}{\shortstack{{\textnormal{[}\textit{atypical}\textnormal{]}} \\-theatre}} 
& FPR95 ↓ & \textbf{58.83} & 67.08 & \underline{63.72} & 69.30 & $64.73 \pm 4.55$ \\
& AUROC ↑ & \textbf{80.94} & 70.68 & \underline{74.13} & 67.49 & $73.31 \pm 5.76$ \\
& AUPR ↑  & \textbf{51.23} & 26.80 & \underline{34.04} & 23.60 & $33.92 \pm 12.34 $\\
\hline
\end{tabular}
}
\label{tab:ood3}
\end{table}



\bibliography{egbib}

\end{document}